\documentclass{bmvc2k}

\usepackage{pbox}
\usepackage{mathtools}

\title{Convolutional Mean: A Simple Convolutional Neural Network for Illuminant Estimation}

\addauthor{Han Gong}{http://www2.cmp.uea.ac.uk/~ybb15eau/}{1}

\addinstitution{
 School of Computing Sciences\\
 University of East Anglia\\
 Norwich, UK
}

\runninghead{Gong}{Convolutional Mean for Simple Illuminant Estimation}

\def\eg{\emph{e.g}\bmvaOneDot}
\def\ie{\emph{i.e}\bmvaOneDot}

\def\etal{\emph{et al}\bmvaOneDot}
\DeclarePairedDelimiter{\norm}{\lVert}{\rVert}

\begin{document}

\maketitle

\begin{abstract}
We present Convolutional Mean (CM) -- a simple and fast convolutional neural network for illuminant estimation. Our proposed method only requires a small neural network model (1.1K parameters) and a $48 \times32$ thumbnail input image. Our unoptimized Python implementation takes 1 ms/image, which is arguably 3-3750$\times$ faster than the current leading solutions with similar accuracy. Using two public datasets, we show that our proposed light-weight method offers accuracy comparable to the current leading methods' (which consist of thousands/millions of parameters) across several measures.
\end{abstract}

%-------------------------------------------------------------------------
\section{Introduction}
\label{sec:intro}
In computer vision, estimating the color of the scene illuminant is a fundamental problem which is commonly known as illuminant estimation. The color cast caused by illumination is usually discounted to support color-based computer vision applications such as image recognition~\cite{SWAIN.IJCV,finlayson2019color}, medical image analysis~\cite{ballerini2013color,hemrit2019using} and general scene understanding~\cite{barrow1981computational}. Illuminant estimation is also useful for ``auto white balance'' -- an essential feature of the modern digital camera. Auto white balance produces natural looking photos by removing the color cast from a raw photo which looks dark and greenish. There have been several hand-crafted methods/features and recent neural network based approaches to tackle this problem. Some of them are simple and efficient however lack accuracy. Other convolution-based methods (\eg \cite{shi2016deep,CCC,bianco2017single} and advanced statistics based methods (\eg \cite{grayedge,finlayson2004shades,CCM}) can achieve better accuracy but they are not fast enough and thus not immediately useful for industrial applications. We conclude that there are mainly three main requirements for deploying an illuminant estimation algorithm to embedded platforms:\\
\textbf{Processing speed} The bundle of all the algorithms running on a digital camera should run at least 30 FPS (frames per second), esp. for video recording or real-time preview. Besides many other tasks, such as object/face detection and artistic image filters, the simple task of illuminant estimation should not take more than 10\% of the total computational time which is about 5 milliseconds per frame~\cite{FFCC}.\\
\textbf{Initialization time} Users would not prefer loading delay when turning on a camera. A practical learning-based illuminant estimation model should contain only a small number of parameters so loading can be instant.\\
\textbf{Thumbnail input} Higher-resolution images are required by most illuminant estimation algorithms for good estimation accuracy. However, processing such large images is costly and impractical for real-time usages. In practice, 8-bit thumbnail images (\eg $48 \times32$ pixels) are usually desirable~\cite{FFCC}.

In this paper, we propose a simple, but effective illuminant estimation algorithm, which is named ``Convolutional Mean'' (CM). CM addresses the above mentioned practical requirements. We see this as an alternative for application scenarios whereby processing speed is prioritized. CM is a small and fast convolutional neural network which offers comparable estimation accuracy on thumbnail input images. Our unoptimized python implementation processes images at 1 milliseconds per image -- arguably 3$\times$faster than FFCC~\cite{FFCC} and 250-4000$\times$ faster than the current leading methods~\cite{shi2016deep,CCC,cheng2015effective}. These features would make CM particularly suitable for embedded deployment (\eg smartphones).

\begin{figure}[htb!]
\includegraphics[width=\linewidth]{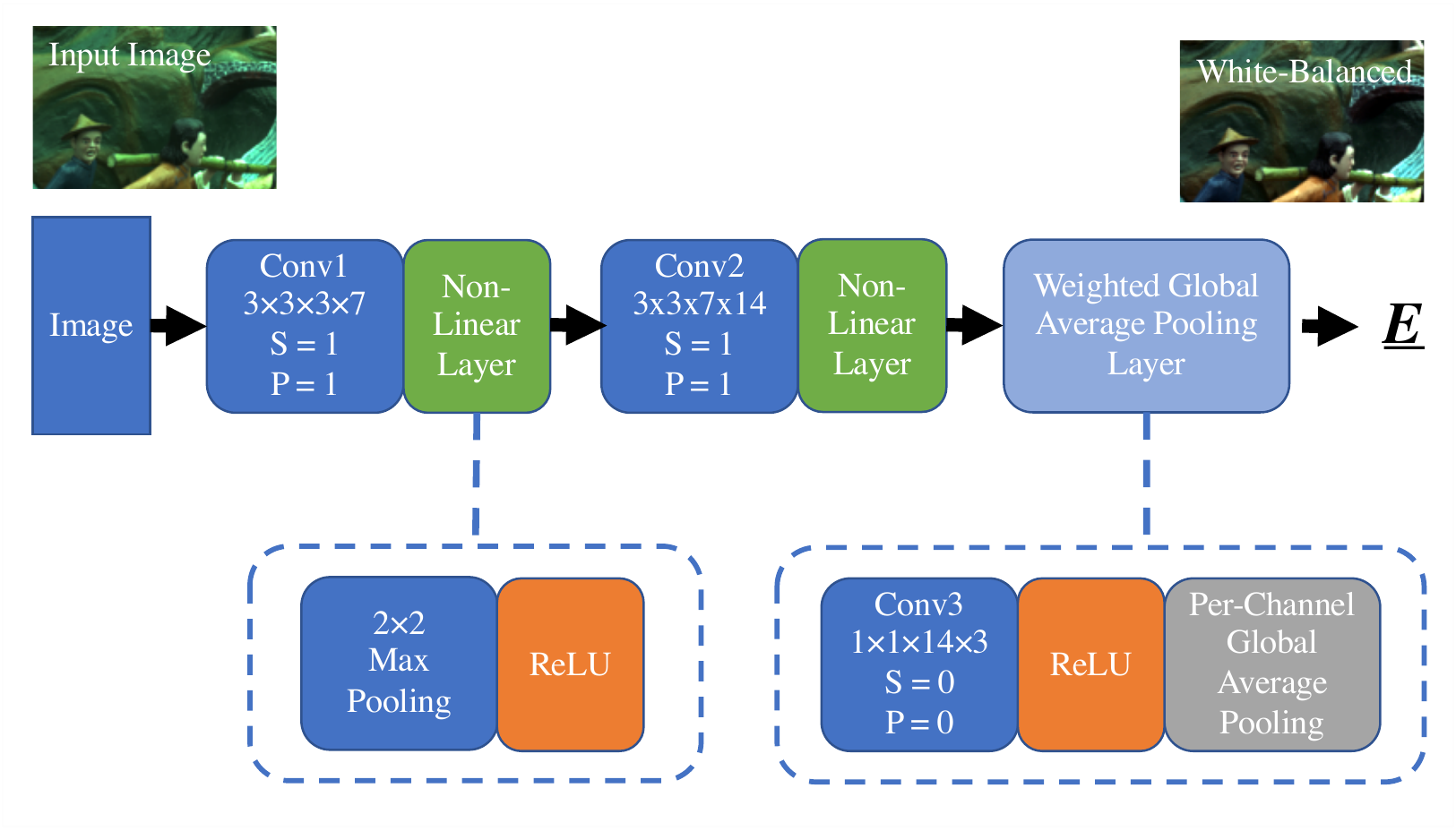}
\caption{Convolutional Mean (CM) network architecture. CM contains two $3\times 3$ filter convolutional layers (Conv1/2) which are followed by a $2\times2$ max pooling and a ReLU. At the end, there is a weighted global averaging layer which is implemented as an $1\times 1$ convolutional layer (Conv3) with ReLU and per-channel global average pooling. In this diagram, $P$ and $S$ denote padding and stride respectively. The other four numbers shown in the Conv blocks represent "Filter Size 1 $\times$ Filter Size 2 $\times$ \#Input Channel $\times$ \#Output Channel" whose product is the total number of filter parameters.}
\label{fig:arch}
\end{figure}

The design of CM (depicted in Figure~\ref{fig:arch}) is surprisingly simple and is inspired by the famous gray world~\cite{grayworld} and gray edge~\cite{grayedge} illuminant estimation algorithms. The design can be briefly summarized as two convolutional layers followed by a weighted per-channel global average pooling layer making use of the mean of all input intensities. Compared with the traditional methods such as gray edge~\cite{grayedge}, our features are not hand-crafted but learned from data. This nature allows for more accurate illuminant estimation at a higher processing speed.

In Section~\ref{sec:rw}, we review the related work on illuminant estimation based on hand-crafted and machine-coded features. In Section~\ref{sec:cm}, we present our new algorithm design and show how to train a light-weight neural network for illuminant estimation. Experiments are presented in Section~\ref{sec:exp}. The paper concludes in Section~\ref{sec:con}.

\section{Related work}
\label{sec:rw}
There have been a lot of literature on illuminant estimation. These methods can be roughly summarized into two categories: (1) Methods based on hand-crafted features. These methods estimate the illuminant by using image statistics or physics assumptions. They include mappings between colors statistics (\eg ~\cite{grayworld,grayedge,chakrabarti2015color}) and bias-correction~\cite{CHCC,CCM,apap}, biologically inspired features (\eg \cite{gijsenij2011color,gijsenij2012improving}), spatial and frequency-domain features from the image and scene illuminations~\cite{bianco2010automatic,CCNATURAL,SpatialCC}, and specularity/shading~\cite{white_patch,finlayson2004shades}. Some of these methods (\eg gray world~\cite{grayworld}) are based on computationally cheap features offering great computational efficiency. However, they generally lack accuracy. The others rely on more advanced features which improve accuracy but at the cost of a lot more computational resources (usually for pre-processing); (2) Methods based on machine-coded features. Given a labelled illuminant ground-truth dataset, researchers train machine learning models for illuminant estimation using supervision. Machine-coded features generally require considerably more encoding parameters and can provide significantly better accuracy compared with hand-crafted features. However, if not handled properly, methods based on machine-coded features would risk over-fitting that the trained models would only work well for the similar data which they have ``seen''. A large-size model can also incur a considerable computational cost (\eg model loading time and processing time) which makes it unsuitable for real-world deployment. As our proposed method also falls into this category, we particularly review some methods based on machine-coded features in the following paragraphs.

The majority of machine-coded features for illuminant estimation have been learned using Convolutional Neural Networks (CNN) which has achieved great success in many computer vision tasks, \eg object recognition~\cite{dosovitskiy2015flownet} and optical flow estimation~\cite{dosovitskiy2015flownet}. Bianco \etal~\cite{bianco2015color} first attempted to adopt a CNN for illuminant estimation which consists of some convolutional layers followed by two fully-connected linear layers. Although the model is heavy and its performance is in fact not better than many methods based on hand-crafted features (\eg~\cite{cheng2015effective}), it has shown potential to adopt CNN for illuminant estimation. This attempt has been followed by recent convolution-based methods which provide substantially improved accuracy. Similar to an earlier Apple patent proposed by Hubel~\etal~\cite{hubel2007white}, Barron~\cite{CCC} has shown that, in the space of 2-D log-chromaticity, convolutional filters can be learned for more accurate illuminant estimation. In his work, illuminant color is re-formulated as a global 2-D translation in the log-chromaticity space. Barron and Tsai~\cite{FFCC} later extended~\cite{CCC} by using FFTs (Fast Fourier Transform) to perform the convolution that filters the log-chromaticity histogram. This method named FFCC is not always more accurate than Barron's previous method -- ``Convolutional Color Constancy'' (CCC)~\cite{CCC} -- but its processing speed is significantly improved. However, both of Barron's methods require a pre-processing step of histogram generation which can be costly. Shi \etal~\cite{shi2016deep} proposed a branch-level ensemble of neural networks consisting of two interacting sub-networks, \ie a hypotheses network and a selection network. The selection network picks for confident estimations from the plausible illuminant estimations generated from the hypotheses network. Shi's method produces accurate results however the model size is huge and its processing speed is slow. The most relevant work to this paper is a confidence-weighted pooling method (named FC4) which is proposed by Hu~\etal~\cite{fc4}. They adopted transfer-learning to train a deep neural network which estimates a per-sub-region illuminant map and a weight map for each sub-region. The illuminant color is the global mean of the weighted per-pixel product between the illuminant map and its weight map. They have achieved some competitive results however their model is huge and significantly slower than FFCC~\cite{FFCC}.

\section{Illuminant Estimation by Convolutional Mean}
\label{sec:cm}
Assuming that an image $I$ is captured by a linear RGB color camera with dark current and saturated pixels removed, the channel $c$ ($c \in \{R,G,B\}$) intensity $I_c$ for a Lambertian surface at pixel $\underline{x}$ can be formulated as the integral of the product of the illuminant spectral power distribution $E(\underline{x},\lambda)$, the surface reflectance $S(\underline{x},\lambda)$ and the sensor response function $Q_c(\lambda)$:

\begin{equation}
I_c(\underline{x}) = \int_{\Omega}{E(\underline{x},\lambda)S(\underline{x},\lambda)Q_c(\underline{x},\lambda)}d\lambda
\label{eq:img_formation}
\end{equation}
where $\lambda$ is the wavelength and $\Omega$ is the visible spectrum. According to the Von Kries coefficient law \cite{white_patch}, Equation~\ref{eq:img_formation} can be simplified as: 
\begin{equation}
I_c(\underline{x}) = E_c \times R_c(\underline{x})
\end{equation}
where $\underline{E}$ is the RGB illumination and $\underline{R}$ is the RGB intensity of reflectance under pure white illumination. For the task of single illuminant estimation, the goal is solving for the global 3-vector illuminant $\underline{E}$.

In this section, we present Convolutional Mean (CM) for illuminant estimation. Our proposed network is a fast and light-weight CNN-based solution. It directly accepts an 8-bit $48\times 32$ thumbnail input image without any significant pre-processing, \eg histogram generation (adopted in \cite{CCC,FFCC}) or homogeneous log-chromaticity intensity conversion (adopted in \cite{shi2016deep,CCC,FFCC}). For industrial applications, our proposed network provides an excellent balance between accuracy and processing/initialization speed.

\subsection{Convolutional Mean}
Our network design is inspired by gray-world~\cite{grayworld} and gray-edge~\cite{grayedge} which assume that the average RGB intensity or edge difference in a scene is achromatic. Their major issue is that not all pixels in an image are useful for illuminant estimation. Despite this significant limitation, gray-world~\cite{grayworld} has gained great popularity because of its low computational cost. In this paper, we attempt to improve this average achromatic intensity idea. Our hypothesis is that through training some shallow non-linear convolutional filters, we could generate selective features for illuminant estimation by simple per-channel global average pooling. The additional non-linearity is introduced by the ReLU and max pooling operators.

Our simple neural network only consists of two convolutional filter layers and a per-channel weighted global average pooling layer using the means of the intermediate outputs produced by the previous convolutional layer. Figure~\ref{fig:arch} shows the detailed network architecture of our proposed CM structure. In the figure, Conv1 and Conv2 generate the machine-coded features for illuminant estimation which are further ``selected'' by the Max-Pooling + ReLU operators. In the last layer of ``weighted per-channel global average pooling'', we first weight each output feature channel after Conv2 (\eg see Figure~\ref{fig:feat}) by using a $1\times 1$ convolutional filter (followed by a ReLU) and obtain a 3-channel output (each channel respectively denotes R, G and B). Finally, we perform the ``gray world'' operation~\cite{grayworld} -- per-channel global average pooling. Mathematically, we can represent our network $f()$ as follows:
\begin{align}
g(I) &= \mathbf{ReLU}(\mathbf{MaxPool}_{2\times2}(I)) \label{eq:g} \\
h(I) &= \mathbf{GW}(\mathbf{ReLU}(I*F^3_{1\times1\times14\times3})) \label{eq:h}\\
f(I) &= h( g( g(I * F^1_{3\times3\times3\times7}) * F^2_{3\times3\times7\times14}) ) \label{eq:f}
\end{align}
where $I$ denotes a multi-channel input array (\eg for $f()$, it denotes a 3-channel input image), $g()$ is a non-linear function formed by a $2\times2$ kernel Max-Pooling and a ReLU, $h()$ is the non-linear weighted averaging function described above, $\mathbf{GW}$ denotes the ``gray-world'' per-channel averaging, $*$ denotes a convolution operation (without a bias term) followed by a set of kernels (\eg $F^{1-3}$ whose subscripts follow the same definition  described in Figure~\ref{fig:arch}).
Note that the resulting 3-vector estimation $\underline{E}$ is up to a scale (which could be linked to exposure difference) and therefore we normalize $\underline{E}$ by dividing its L2-norm. As shown in Figure~\ref{fig:arch}, our total number of parameter is $1,113$. By default, we also normalize the intensities of $I$ by dividing by its global maximum intensity -- a scalar.

\subsection{Network training}
\begin{figure}[htb!]
\includegraphics[width=\linewidth]{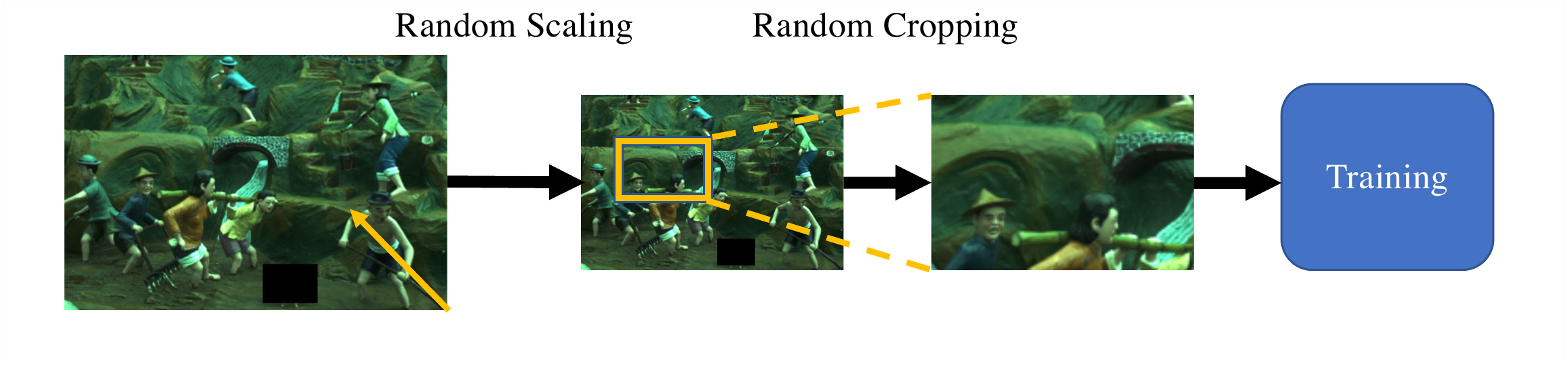}
\caption{Training patch cropping procedure. The yellow frame indicates the cropped patch.}
\label{fig:crop}
\end{figure}

We have adopted the same training image datasets used by FFCC~\cite{FFCC} and CCC~\cite{CCC}. In their pre-processed datasets, all the image regions belonging to the color/gray checkers have been masked out (wiped as black -- $0$ intensity). A common limitation in these datasets is that their numbers of samples are too small relative to the number of model parameters required. Therefore, data augmentation is required for training. Although our neural network works for images in different resolutions, we still specify a standard working resolution of $48\times 32$. Given a higher-resolution $384 \times 256$ training image, we first randomly re-size it to a scale between $0.125$ to $1$ of the original size (using bi-linear interpolation). Then, we randomly crop a $48\times 32$ (\ie standard working resolution) image patch from the previously re-scaled image. This cropping step finalizes the pre-processing for training. Note that these pre-processing steps of data augmentation are not required for execution. Figure~\ref{fig:crop} shows an example of this procedure.

In the training phase, we adopt the popular optimization algorithm -- Adam~\cite{adam} -- using the following settings: 1) learning rate = $10^{-3}$; 2) batch size = $16$; 3) number of epoch = 2000; 4) loss function = L1-norm. We have tried the other loss functions such as L2-norm or angular error. In practice, L1-norm gives the best results; 5) weight initialization: Kaiming normal distribution~\cite{he2015delving}. The training dataset is too small and the additional dataset slicing for testing would not be practical. To avoid over-fitting, we still require a test set that in each epoch we test the accuracy of the trained model. Since only the small randomly cropped thumbnail-size patches are used for training, the thumbnail version of all the \textit{uncropped} training images are visually different from the cropped images and they have been adopted as the test set to compute the test error for the trained model of each epoch. The model which produces the minimum test error is then selected as the final model (\eg for 3-fold cross-validation). In summary, given a higher-resolution training image, we have used its randomly cropped thumbnail-size patches for training and its uncropped thumbnail image for testing. There is no overlap between training/testing images and validation images. We will show that this tactic is effective in the following section of evaluation.

\section{Evaluation}
\label{sec:exp}
% per error output (final layer), two datasets, higher-res image input.

We have implemented our neural network using PyTorch~\cite{pytorch}. Following the similar evaluation carried out in FFCC~\cite{FFCC}, we evaluate our method -- CM -- using two popular color constancy datasets: the NUS dataset~\cite{cheng2014illuminant} and the Gehler-Shi dataset~\cite{gehler2008bayesian} reprocessed by Shi and Funt~\cite{shi_reprocessed}. We adopt 3-fold cross-validation for our evaluation. Note that all the measurements are calculated using the concatenated errors of the three folds.
\subsection{Experiment results and discussions}
The results are shown in Tables \ref{tbl:res2} and \ref{tbl:res1} where angular error is used to report the results. Angular error $e$ is defined as follows:
\begin{equation}
e = \mathrm{acos} \bigg( \frac{\underline{E} \cdot \underline{E}_\text{gt}}{\norm{\underline{E}} \norm{\underline{E}_{\text{gt}}} } \bigg)
\end{equation}
where $\underline{E}_\text{gt}$ denotes the illuminant ground truth -- a 3-vector, $\norm{.}$ denotes an L2 norm.

In our evaluation, we focus on the processing accuracy for thumbnail resolution ($48 \times 32$) 8-bit images which are practical for deploying a white-balance system on embedded devices. The evaluation results of the other listed methods are based on their recommended image resolutions and bit depth reported in the corresponding papers. As seen in Table~\ref{tbl:res2}, our illuminant estimation accuracy (esp. for mean and median) is close to the leading methods such as \cite{FFCC,CCC,cheng2015effective,shi2016deep,fc4} and the overall results in Table~\ref{tbl:res1} are somewhat worse than the leading methods~\cite{FFCC,CCC,shi2016deep,fc4}. It is worth noting that FFCC~\cite{FFCC} and our CM only take 8-bit $48 \times 32$ (thumbnail) resolution input images while the others take 16-bit original resolution input images. Our CM is also end-to-end without requiring any pre-processing (\eg histogram generation used in FFCC~\cite{FFCC} or transferred feature extractor~\cite{fc4}).

\begin{table}[htb!]
\begin{center}
\begin{tabular}{|l|ccccc|}
\hline
Algorithm & Mean & Med. & Tri. & Best & Worst \\
&&&& 25\% & 25\%\\
\hline\hline
White-Patch~\cite{white_patch} & 9.91 & 7.44 & 8.78 & 1.44 & 21.27 \\
Pixels-based Gamut~\cite{gijsenij2010generalized} & 5.27 & 4.26 & 4.45 & 1.28 & 11.16  \\
Grey-world~\cite{grayworld} & 4.59 & 3.46 & 3.81 & 1.16 & 9.85 \\
Edge-based Gamut~\cite{gijsenij2010generalized} & 4.40 & 3.30 & 3.45 & 0.99& 9.83  \\
Shades-of-Gray~\cite{finlayson2004shades} & 3.67 & 2.94 & 3.03 & 0.98 & 7.75  \\
Bayesian~\cite{gehler2008bayesian} & 3.50 & 2.36 & 2.57 & 0.78 & 8.02  \\
Natural Image Statistics~\cite{gijsenij2011color} & 3.45 & 2.88 & 2.95 & 0.83 & 7.18  \\
LSRS~\cite{gao2014efficient} & 3.45 & 2.51 & 2.70 & 0.98 & 7.32 \\
2nd-order Gray-Edge~\cite{grayedge} & 3.36 & 2.70 & 2.80 & 0.89 & 7.14  \\
1st-order Gray-Edge~\cite{grayedge} & 3.35 & 2.58 & 2.76 & 0.79 & 7.18  \\
General Gray-World~\cite{barnard2002comparison} & 3.20 & 2.56 & 2.68 & 0.85 & 6.68 \\ 
Spatio-Spectral Statistics~\cite{SpatialCC} & 3.06 & 2.58 & 2.74 & 0.87 & 6.17 \\
Corrected-Moment~\cite{CCM} & 2.95 & 2.05 & 2.16 & 0.59 & 6.89 \\
Bright-and-Dark Colors PCA~\cite{cheng2014illuminant} & 2.93 & 2.33 & 2.42 & 0.78 & 6.13  \\
Color Dog~\cite{colordog} & 2.83 & 1.77 & 2.03 & 0.48 & 7.04  \\
Homography~\cite{CHCC} & 2.55 & 1.70 & - & - & 5.78 \\
APAP-LUT~\cite{apap} (GW) & 2.52 & 1.83 & - & 0.60 & 5.62 \\
CCC~\cite{CCC} & 2.38 & 1.48 & 1.69 & 0.45 & 5.85 \\
Deep Specialized Net~\cite{shi2016deep} & 2.24 & 1.46 & 1.68 & 0.48 & 6.08 \\ 
Regression Tree~\cite{cheng2015effective} & 2.18 & 1.48 & 1.64 & 0.46 & 5.03 \\
FC4~\cite{fc4} (AlexNet) & 2.12 & 1.53 & 1.67 & 0.48 & 6.08 \\
FFCC~\cite{FFCC} (Model Q) & \textbf{2.06} & \textbf{1.39} & \textbf{1.53} & \textbf{0.39} & \textbf{4.80} \\
\hline
CM (Proposed) & 2.25 & 1.59 & 1.74 & 0.50 & 5.13\\
\hline
\end{tabular}
\end{center}
\caption{Performance on the dataset from Cheng et al.~\cite{cheng2014illuminant}. We present five error metrics ranked by mean error. As was shown in \cite{CCC,FFCC}, we present the average performance (the geometric mean) over all 8 cameras in the dataset. The best scores are made bold. ``Tri.'' and ``Med.'' refer to Trimean and Median respectively.}
\label{tbl:res2}
\end{table}

Our method requires fewer model parameters and it provides a leading balance between accuracy and speed. As for model size, we show a comparison with some leading methods in Table~\ref{tbl:res1}. Our model parameter size is $157\%$ of CCC~\cite{CCC}, $14\%$ of FFCC~\cite{FFCC}, $0.021\%$ of Deep Specialized Network~\cite{shi2016deep}, $0.025\%$ of FC4~\cite{fc4}, and $0.00003\%$ of Regression Tree~\cite{cheng2015effective}. Note that although CCC~\cite{CCC} requires fewer parameters, it is significantly slower than ours. The number of model parameters affects the initialization time of the imaging system (\eg for loading parameters to memory). Assuming that we adopt 32-bit floating numbers for storing our model parameters, the initialization of our model would require loading $4.4$ KB data which is fairly light (\ie an unnoticeable delay).
In terms of processing speed, our unoptimized python implementation takes $1$ms (on a Tesla K40m GPU) to process an image which is 3$\times$ faster than the unoptimized FFCC~\cite{FFCC} ($2.37$ ms/image), 312$\times$ faster than Regression Tree~\cite{cheng2015effective} ($0.25$s/image), 650$\times$ faster than CCC~\cite{CCC}, 31$\times$ faster than FC4~\cite{fc4} (GPU) and 3750$\times$ faster than Deep Specialized Network~\cite{shi2016deep} (GPU). Note that this speed comparison is based on modern PC platforms for all methods and the fine-grained CPU/GPU differences are not considered. However, given the much simpler model and the faster speed compared with the unoptimized PC version of FFCC~\cite{FFCC}, we believe that our CM can arguably take less than 5\% computational budget to support at least a 30-60 FPS embedded imaging system (estimated according to the optimized performance of FFCC~\cite{FFCC}). This computational efficiency would be desirable for embedded deployment. We remark that future rigorous tests are still required for comparing the actual performance on embedded platforms.

\begin{table}[htb!]
\begin{center}
\begin{tabular}{|l|ccccc|cc|}
\hline
Algorithm & Mean & Med. & Tri. & Best & Worst & Test & Para. \\
&&&& 25\% & 25\% & Time & No.\\
\hline\hline
SVR~\cite{funt2004estimating} & 8.08 & 6.73 & 7.19 & 3.35 & 14.89 & - & - \\
White-Patch~\cite{white_patch} & 7.55 & 5.68 & 6.35 & 1.45 & 16.12 & 0.16 & - \\
Grey-World~\cite{grayworld} & 6.36 & 6.28 & 6.28 & 2.33 & 10.58 & 0.15 & - \\
1st-Order Gray-Edge~\cite{grayedge} & 5.33 & 4.52 & 4.73 & 1.86 & 10.03 & 1.1 & - \\
2nd-Order Gray-Edge~\cite{grayedge} & 5.13 & 4.44 & 4.62 & 2.11 & 9.26 & 1.3 & - \\
Shades-of-Gray~\cite{finlayson2004shades} & 4.93 & 4.01 & 4.23 & 1.14 & 10.20 & 0.47 & - \\
Bayesian~\cite{gehler2008bayesian} & 4.82 & 3.46 & 3.88 & 1.26 & 10.49 & 97 & - \\
Yang et al. 2015~\cite{yang2015efficient} & 4.60 & 3.10 & - & - & - & 0.88 & - \\
General Gray-World~\cite{barnard2002comparison}& 4.66 & 3.48 & 3.81 & 1.00 & 10.09 & 0.91 & - \\
Natural Image Statistics~\cite{gijsenij2011color}& 4.19 & 3.13 & 3.45 & 1.00 & 9.22 & 1.5 & - \\
CART-Based Combination~\cite{bianco2010automatic}& 3.90 & 2.91 & 3.21 & 1.02 & 8.27 & - & - \\
Spatio-Spectral Statistics~\cite{SpatialCC} & 3.59 & 2.96 & 3.10 & 0.95 & 7.61 & 6.9 & - \\
LSRS~\cite{gao2014efficient} & 3.31 & 2.80 & 2.87 & 1.14 & 6.39 & 2.6 & - \\
Pixels-Based Gamut~\cite{gijsenij2010generalized} & 4.20 & 2.33 & 2.91 & 0.50 & 10.72 & - & - \\
Bottom-up+Top-down~\cite{van2007using} & 3.48 & 2.47 & 2.61 & 0.84 & 8.01 & - & - \\
Cheng et al. 2014~\cite{cheng2014illuminant} & 3.52 & 2.14 & 2.47 & 0.50 & 8.74 & 0.24 & - \\
Exemplar-based~\cite{EXEMPLAR} & 2.89 & 2.27 & 2.42 & 0.82 & 5.97 & - & - \\
Bianco et al. 2015~\cite{bianco2015color} & 2.63 & 1.98 & - & - & - & - & 0.15M\\
APAP-LUT~\cite{apap} (GW) & 2.96 & 2.22 & - & 0.59 & 6.58 & 0.011 & 256\\
Corrected-Moment~\cite{CCM} & 2.86 & 2.04 & 2.22 & 0.70 & 6.34 & 0.77 & 57 \\
Charkrabarti et al. 2015~\cite{chakrabarti2015color} & 2.56 & 1.67 & 1.89 & 0.52 & 6.07 & 0.30 & - \\
Regression Tree~\cite{cheng2015effective} & 2.42 & 1.65 & 1.75 & 0.38 & 5.87 & 0.25 & 31.5M \\
FFCC~\cite{FFCC} (Model Q) & 2.01 & 1.13 & 1.38 & \textbf{0.30} & 5.14 & 0.0024 & 8.2K \\
CCC~\cite{CCC} & 1.95 & 1.22 & 1.38 & 0.35 & 4.76 & 0.52 & 0.7K \\
Deep Specialized Net~\cite{shi2016deep} & 1.90 & 1.12 & 1.33 & 0.31 & 4.84 & 3 & 5.3M \\
FC4~\cite{fc4} (AlexNet)& \textbf{1.77} & \textbf{1.11} & \textbf{1.29} & 0.34 & \textbf{4.29} & 0.025 & 4.34M \\
\hline
CM (Proposed) & 2.48 & 1.61 & 1.80 & 0.47 & 5.97 & \textbf{0.001} & 1.1K \\
\hline
CM-A (Without $\mathbf{MaxPool}$) & 2.56 & 1.70 & 1.87 & 0.48 & 6.15 & 0.001 & 1.1K \\
CM-B (Without $\mathbf{ReLU}$) & 2.66 & 1.79 & 1.96 & 0.51 & 6.34 & 0.001 & 1.1K \\
CM-C (Single Conv. Layer) & 2.49 & 1.67 & 1.83 & 0.50 & 5.87 & 0.001 & 1.1K \\
CM-D (rgb Chroma. Input) & 3.03 & 2.14 & 2.34 & 0.68 & 6.90 & 0.001 & 1.1K \\
CM-E (Without a Test Set) & 2.62 & 1.73 & 1.91 & 0.49 & 6.30 & 0.001 & 1.1K \\
\hline
\end{tabular}
\end{center}
\caption{Performance on the Gehler-Shi dataset~\cite{gehler2008bayesian,shi_reprocessed} in the same format as Table~\ref{tbl:res2}. We present the test time (in seconds) for evaluating a single image, when available. The best scores are made bold. K and M denote thousand and million respectively. ``Tri.'' and ``Med.'' refer to Trimean and Median respectively.}
\label{tbl:res1}
\end{table}

%In addition, using the same model trained with 8-bit thumbnail images, we have tested its illuminant estimation accuracy given full-resolution ($384 \times 256$) 16-bit input images. In this case, we have observed a similar result. This additional experiment shows that our trained models could be compatible with higher image resolutions which is not the case for some other CNN-based solutions such as \cite{shi2016deep,bianco2017single}. However, from a practical point of view, the ability to deal with thumbnail 8-bit images using least modelling parameters and computational resources would still be the most focus for real-world applications.

As for its variants, we have tried the following options based on the Gehler-Shi  dataset~\cite{gehler2008bayesian,shi_reprocessed} (listed in Table~\ref{tbl:res1}):\\
A) \textit{Without ReLU}. The overall results are worse;
B) \textit{Without max pooling}. The overall results are worse;
C) \textit{Single convolutional layer}. We use the similar number of parameters  however they are assigned to a single convolutional layer with more channels (38 channels) that Equations \ref{eq:h} and \ref{eq:f} are replaced with the follows:
\begin{align}
h(I) &= \mathbf{GW}(\mathbf{ReLU}(I*F^3_{1\times1\times38\times3})) \\
f(I) &= h( g(I * F^1_{3\times3\times3\times38}) )
\label{eq:net1}
\end{align}
The results are worse in all the measures. We did not attempt to make our network deeper than two convolutional layers as deeper networks would be more difficult to train and are not necessarily more efficient for illuminant estimation compared with simpler structures;
D) \textit{rgb chromaticity input}. Instead of using RGB input images, we convert the RGBs to their rgb chromaticities. However, the results are significantly worse. This could be caused by the loss of shading information which has been used as an important cue for some previous methods (\eg gray edge~\cite{grayedge}).
Through the test of CM variants, we can conclude that the introduced non-linearity, additional depth, and the preserved shading information are helpful for improving illuminant estimation accuracy;
E) \textit{Without a test set}. We found that CM tends to over-fit (\ie poorer accuracy) when the test set images -- uncropped thumbnail images -- are not used in training.

\subsection{Learned Knowledge}
\begin{figure}[htb!]
\includegraphics[width=\linewidth]{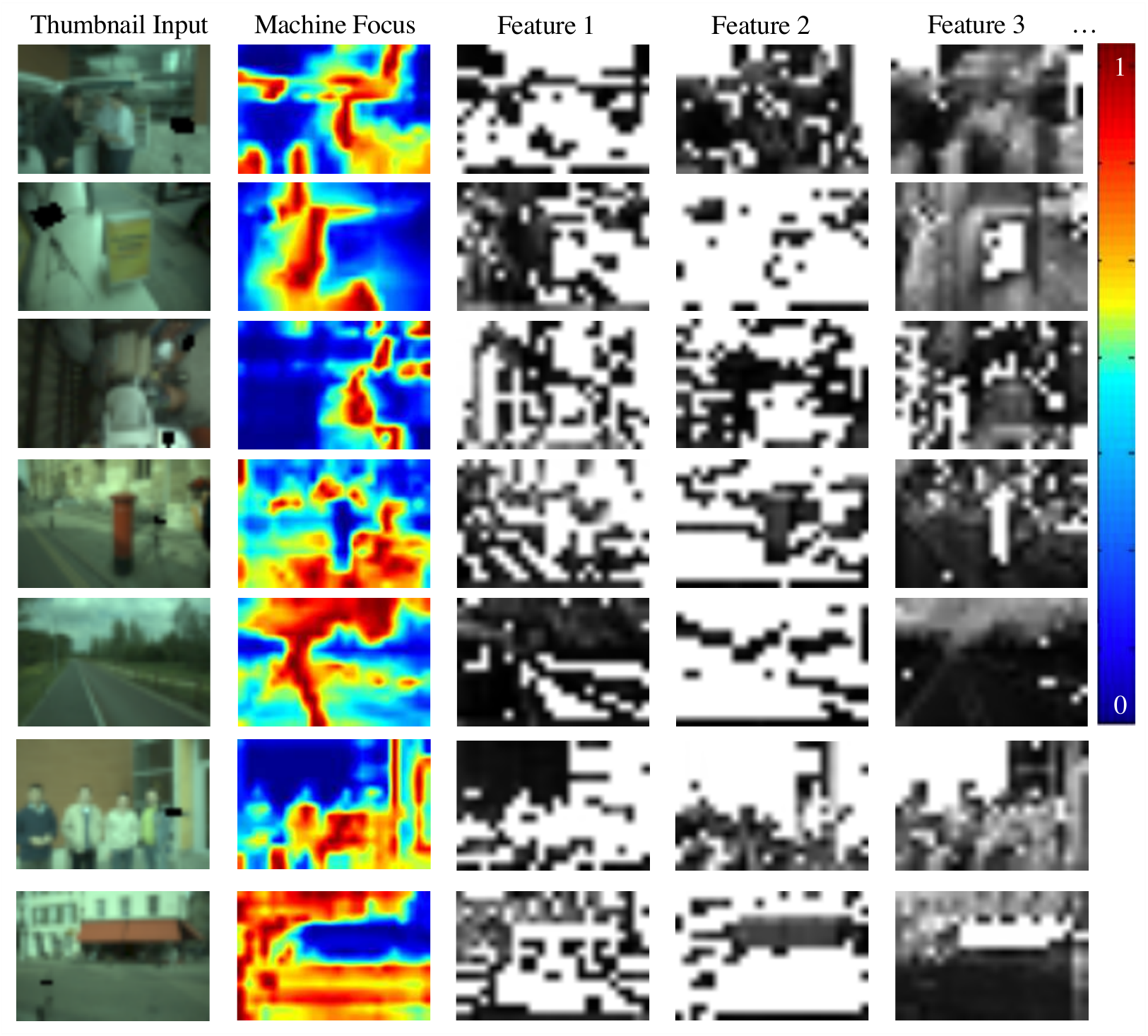}
\caption{Features learned by our CM. The first column is the original input image. In the last three columns, we show the first three channels of the learned features. In the second column, we show the normalized machine-focus map wherein the reddish pixels indicate the areas which contribute more to illuminant estimation. All the images are up-sampled for visualization.}
\label{fig:feat}
\end{figure}

Since the final weighted per-channel global average pooling layer is essentially a fusion of all filtered image features, visualizing these filtered image features would be helpful to understand what has been learned. In Figure~\ref{fig:feat}, given some inputs, we visualize the first 3 (of 14) channels of the learned intermediate features. We have observed both sparse features and smooth features, \eg Feature 3 looks relevant to colorfulness.

Since the final output is computed by per-channel averaging the last 3-channel network responses (after Conv2), most of the filtered pixel intensities should be close to the illuminant ground truth and the brighter pixels should contribute more to the final estimate. We convert the last 3-channel response image to a gray-scale image by taking a channel-wise average. In this gray-scale image, the brighter regions are more focused by our trained model for illuminant estimation. Some of these examples are shown in Figure~\ref{fig:feat}. The trained model seems to focus on grayer surfaces for illuminant estimation. This pixel selectivity which CM offers is one of the fundamental differences compared with gray world~\cite{grayworld} and gray edge~\cite{grayedge}.

%-------------------------------------------------------------------------
\section{Conclusion}
\label{sec:con}
We have presented Convolutional Mean (CM) -- a simple and fast algorithm for illuminant estimation. Our proposed method accepts $48 \times32$ thumbnail input images for real-time processing (at least 30-60 frames per seconds with 5\% computational budget) which is arguably 3-4500$\times$ faster than the other leading solutions. We have also shown that our proposed light-weight method offers accuracy comparable to the leading methods' (which are relatively more parameter-demanding) across several measures. Future work would be a further reduction of model parameters, a full performance verification on embedded platforms, and a trail of other efficient statistics combined with machine-coded features.

\section*{Acknowledgements}
The model training was carried out on the High Performance Computing Cluster supported by the Research and Specialist Computing Support service at the University of East Anglia. We thank NVIDIA for their generous donation of a GPU. We also thank the anonymous reviewers for their constructive feedback.

\bibliography{biblio}

\begin{thebibliography}{41}
\providecommand{\natexlab}[1]{#1}
\providecommand{\url}[1]{\texttt{#1}}
\expandafter\ifx\csname urlstyle\endcsname\relax
  \providecommand{\doi}[1]{doi: #1}\else
  \providecommand{\doi}{doi: \begingroup \urlstyle{rm}\Url}\fi

\bibitem[pyt()]{pytorch}
Pytorch.
\newblock URL \url{https://pytorch.org/}.

\bibitem[Afifi et~al.(2019)Afifi, Punnappurath, Finlayson, and Brown]{apap}
Mahmoud Afifi, Abhijith Punnappurath, Graham Finlayson, and Michael~S Brown.
\newblock As-projective-as-possible bias correction for illumination estimation
  algorithms.
\newblock \emph{Journal of the Optical Society of America A}, 36\penalty0
  (1):\penalty0 71--78, 2019.

\bibitem[Ballerini et~al.(2013)Ballerini, Fisher, Aldridge, and
  Rees]{ballerini2013color}
Lucia Ballerini, Robert~B Fisher, Ben Aldridge, and Jonathan Rees.
\newblock A color and texture based hierarchical k-nn approach to the
  classification of non-melanoma skin lesions.
\newblock In \emph{Color Medical Image Analysis}, pages 63--86. Springer, 2013.

\bibitem[Banic and Loncaric(2015)]{colordog}
Nikola Banic and Sven Loncaric.
\newblock Color dog-guiding the global illumination estimation to better
  accuracy.
\newblock In \emph{International Conference on Computer Vision Theory and
  Applications}, pages 129--135, 2015.

\bibitem[Barnard et~al.(2002)Barnard, Cardei, and Funt]{barnard2002comparison}
Kobus Barnard, Vlad Cardei, and Brian Funt.
\newblock A comparison of computational color constancy algorithms. i:
  Methodology and experiments with synthesized data.
\newblock \emph{IEEE transactions on Image Processing}, 11\penalty0
  (9):\penalty0 972--984, 2002.

\bibitem[Barron(2015)]{CCC}
Jonathan~T Barron.
\newblock Convolutional color constancy.
\newblock In \emph{IEEE International Conference on Computer Vision}, pages
  379--387, 2015.

\bibitem[Barron and Tsai(2017)]{FFCC}
Jonathan~T Barron and Yun-Ta Tsai.
\newblock Fast fourier color constancy.
\newblock In \emph{IEEE Conference on Computer Vision and Pattern Recognition},
  pages 886--894, 2017.

\bibitem[Barrow and Tenenbaum(1981)]{barrow1981computational}
Harry~G Barrow and Jay~M Tenenbaum.
\newblock Computational vision.
\newblock \emph{Proceedings of the IEEE}, 69\penalty0 (5):\penalty0 572--595,
  1981.

\bibitem[Bianco et~al.(2010)Bianco, Ciocca, Cusano, and
  Schettini]{bianco2010automatic}
Simone Bianco, Gianluigi Ciocca, Claudio Cusano, and Raimondo Schettini.
\newblock Automatic color constancy algorithm selection and combination.
\newblock \emph{Pattern recognition}, 43\penalty0 (3):\penalty0 695--705, 2010.

\bibitem[Bianco et~al.(2015)Bianco, Cusano, and Schettini]{bianco2015color}
Simone Bianco, Claudio Cusano, and Raimondo Schettini.
\newblock Color constancy using cnns.
\newblock In \emph{IEEE Conference on Computer Vision and Pattern Recognition
  Workshops}, pages 81--89, 2015.

\bibitem[Bianco et~al.(2017)Bianco, Cusano, and Schettini]{bianco2017single}
Simone Bianco, Claudio Cusano, and Raimondo Schettini.
\newblock Single and multiple illuminant estimation using convolutional neural
  networks.
\newblock \emph{IEEE Transactions on Image Processing}, 26\penalty0
  (9):\penalty0 4347--4362, 2017.

\bibitem[Brainard and Wandell(1986)]{white_patch}
David~H Brainard and Brian~A Wandell.
\newblock Analysis of the retinex theory of color vision.
\newblock \emph{Journal of Optical Society of America A}, 3\penalty0
  (10):\penalty0 1651--1661, 1986.

\bibitem[Buchsbaum(1980)]{grayworld}
Gershon Buchsbaum.
\newblock A spatial processor model for object colour perception.
\newblock \emph{Journal of the Franklin institute}, 310\penalty0 (1):\penalty0
  1--26, 1980.

\bibitem[Chakrabarti(2015)]{chakrabarti2015color}
Ayan Chakrabarti.
\newblock Color constancy by learning to predict chromaticity from luminance.
\newblock In \emph{Advances in Neural Information Processing Systems}, pages
  163--171, 2015.

\bibitem[Chakrabarti et~al.(2012)Chakrabarti, Hirakawa, and Zickler]{SpatialCC}
Ayan Chakrabarti, Keigo Hirakawa, and Todd Zickler.
\newblock Color constancy with spatio-spectral statistics.
\newblock \emph{IEEE Transactions on Pattern Analysis and Machine
  Intelligence}, pages 1508--1517, 2012.

\bibitem[Cheng et~al.(2014)Cheng, Prasad, and Brown]{cheng2014illuminant}
Dongliang Cheng, Dilip~K Prasad, and Michael~S Brown.
\newblock Illuminant estimation for color constancy: why spatial-domain methods
  work and the role of the color distribution.
\newblock \emph{JOSA A}, 31\penalty0 (5):\penalty0 1049--1058, 2014.

\bibitem[Cheng et~al.(2015)Cheng, Price, Cohen, and Brown]{cheng2015effective}
Dongliang Cheng, Brian Price, Scott Cohen, and Michael~S Brown.
\newblock Effective learning-based illuminant estimation using simple features.
\newblock In \emph{IEEE Conference on Computer Vision and Pattern Recognition},
  pages 1000--1008, 2015.

\bibitem[Diederik P.~Kingma(2014)]{adam}
Jimmy~Ba Diederik P.~Kingma.
\newblock Adam: A method for stochastic optimization.
\newblock In \emph{International Conference on Learning Representations}, 2014.

\bibitem[Dosovitskiy et~al.(2015)Dosovitskiy, Fischer, Ilg, Hausser, Hazirbas,
  Golkov, Van Der~Smagt, Cremers, and Brox]{dosovitskiy2015flownet}
Alexey Dosovitskiy, Philipp Fischer, Eddy Ilg, Philip Hausser, Caner Hazirbas,
  Vladimir Golkov, Patrick Van Der~Smagt, Daniel Cremers, and Thomas Brox.
\newblock Flownet: Learning optical flow with convolutional networks.
\newblock In \emph{IEEE international conference on computer vision}, pages
  2758--2766, 2015.

\bibitem[Finlayson et~al.(2019)Finlayson, Gong, and Fisher]{finlayson2019color}
Graham Finlayson, Han Gong, and Robert~B Fisher.
\newblock Color homography: theory and applications.
\newblock \emph{IEEE transactions on pattern analysis and machine
  intelligence}, 41\penalty0 (1):\penalty0 20--33, 2019.

\bibitem[Finlayson(2013)]{CCM}
Graham~D Finlayson.
\newblock Corrected-moment illuminant estimation.
\newblock In \emph{International Conference on Computer Vision}, pages
  1904--1911. IEEE, 2013.

\bibitem[Finlayson(2018)]{CHCC}
Graham~D Finlayson.
\newblock Colour and illumination in computer vision.
\newblock \emph{Interface focus}, 8\penalty0 (4):\penalty0 20180008, 2018.

\bibitem[Finlayson and Trezzi(2004)]{finlayson2004shades}
Graham~D Finlayson and Elisabetta Trezzi.
\newblock Shades of gray and colour constancy.
\newblock In \emph{Color and Imaging Conference}, volume 2004, pages 37--41.
  Society for Imaging Science and Technology, 2004.

\bibitem[Funt and Xiong(2004)]{funt2004estimating}
Brian Funt and Weihua Xiong.
\newblock Estimating illumination chromaticity via support vector regression.
\newblock In \emph{Color and Imaging Conference}, volume 2004, pages 47--52.
  Society for Imaging Science and Technology, 2004.

\bibitem[Gao et~al.(2014)Gao, Han, Yang, Li, and Li]{gao2014efficient}
Shaobing Gao, Wangwang Han, Kaifu Yang, Chaoyi Li, and Yongjie Li.
\newblock Efficient color constancy with local surface reflectance statistics.
\newblock In \emph{European Conference on Computer Vision}, pages 158--173.
  Springer, 2014.

\bibitem[Gehler et~al.(2008)Gehler, Rother, Blake, Minka, and
  Sharp]{gehler2008bayesian}
Peter~Vincent Gehler, Carsten Rother, Andrew Blake, Tom Minka, and Toby Sharp.
\newblock Bayesian color constancy revisited.
\newblock In \emph{2008 IEEE Conference on Computer Vision and Pattern
  Recognition}, pages 1--8. IEEE, 2008.

\bibitem[Gijsenij and Gevers(2011{\natexlab{a}})]{CCNATURAL}
Arjan Gijsenij and Theo Gevers.
\newblock Color constancy using natural image statistics and scene semantics.
\newblock \emph{IEEE Transactions Pattern Analysis Machine Intelligence},
  33\penalty0 (4):\penalty0 687--698, 2011{\natexlab{a}}.

\bibitem[Gijsenij and Gevers(2011{\natexlab{b}})]{gijsenij2011color}
Arjan Gijsenij and Theo Gevers.
\newblock Color constancy using natural image statistics and scene semantics.
\newblock \emph{IEEE Transactions on Pattern Analysis and Machine
  Intelligence}, 33\penalty0 (4):\penalty0 687--698, 2011{\natexlab{b}}.

\bibitem[Gijsenij et~al.(2010)Gijsenij, Gevers, and Van
  De~Weijer]{gijsenij2010generalized}
Arjan Gijsenij, Theo Gevers, and Joost Van De~Weijer.
\newblock Generalized gamut mapping using image derivative structures for color
  constancy.
\newblock \emph{International Journal of Computer Vision}, 86\penalty0
  (2-3):\penalty0 127--139, 2010.

\bibitem[Gijsenij et~al.(2012)Gijsenij, Gevers, and Van
  De~Weijer]{gijsenij2012improving}
Arjan Gijsenij, Theo Gevers, and Joost Van De~Weijer.
\newblock Improving color constancy by photometric edge weighting.
\newblock \emph{IEEE Transactions on Pattern Analysis and Machine
  Intelligence}, 34\penalty0 (5):\penalty0 918--929, 2012.

\bibitem[He et~al.(2015)He, Zhang, Ren, and Sun]{he2015delving}
Kaiming He, Xiangyu Zhang, Shaoqing Ren, and Jian Sun.
\newblock Delving deep into rectifiers: Surpassing human-level performance on
  imagenet classification.
\newblock In \emph{IEEE international conference on computer vision}, pages
  1026--1034, 2015.

\bibitem[Hemrit et~al.(2019)Hemrit, Matsushita, Uchida, Vazquez-Corral, Gong,
  Tsumura, and Finlayson]{hemrit2019using}
Ghalia Hemrit, Futa Matsushita, Mihiro Uchida, Javier Vazquez-Corral, Han Gong,
  Norimichi Tsumura, and Graham~D Finlayson.
\newblock Using the monge-kantorovitch transform in chromagenic color constancy
  for pathophysiology.
\newblock In \emph{International Workshop on Computational Color Imaging},
  pages 121--133. Springer, 2019.

\bibitem[Hu et~al.(2017)Hu, Wang, and Lin]{fc4}
Yuanming Hu, Baoyuan Wang, and Stephen Lin.
\newblock Fc4: Fully convolutional color constancy with confidence-weighted
  pooling.
\newblock In \emph{Conference on Computer Vision and Pattern Recognition},
  pages 4085--4094. IEEE, 2017.

\bibitem[Hubel et~al.(2007)Hubel, Finlayson, and Hordley]{hubel2007white}
Paul~M Hubel, Graham~D Finlayson, and Steven~D Hordley.
\newblock White point estimation using color by convolution, April~3 2007.
\newblock US Patent 7,200,264.

\bibitem[Joze and Drew(2012)]{EXEMPLAR}
Hamid Reza~Vaezi Joze and Mark Drew.
\newblock Exemplar-based colour constancy.
\newblock In \emph{British Machine Vision Conference}, pages 26.1--26.12. BMVA
  Press, 2012.

\bibitem[Shi and Funt()]{shi_reprocessed}
L.~Shi and B.~Funt.
\newblock Re-processed version of the gehler color constancy dataset of 568
  images.
\newblock URL \url{http://www.cs.sfu.ca/~colour/data/shi_gehler/}.

\bibitem[Shi et~al.(2016)Shi, Loy, and Tang]{shi2016deep}
Wu~Shi, Chen~Change Loy, and Xiaoou Tang.
\newblock Deep specialized network for illuminant estimation.
\newblock In \emph{European Conference on Computer Vision}, pages 371--387.
  Springer, 2016.

\bibitem[{S}wain and {B}allard(1991)]{SWAIN.IJCV}
{M.J}. {S}wain and {D.H.}. {B}allard.
\newblock {C}olor indexing.
\newblock \emph{{I}nternational {J}ournal of {C}omputer {V}ision}, 7\penalty0
  (11):\penalty0 11--32, 1991.

\bibitem[Van De~Weijer et~al.(2007{\natexlab{a}})Van De~Weijer, Gevers, and
  Gijsenij]{grayedge}
Joost Van De~Weijer, Theo Gevers, and Arjan Gijsenij.
\newblock Edge-based color constancy.
\newblock \emph{IEEE Transactions on image processing}, 16\penalty0
  (9):\penalty0 2207--2214, 2007{\natexlab{a}}.

\bibitem[Van De~Weijer et~al.(2007{\natexlab{b}})Van De~Weijer, Schmid, and
  Verbeek]{van2007using}
Joost Van De~Weijer, Cordelia Schmid, and Jakob Verbeek.
\newblock Using high-level visual information for color constancy.
\newblock In \emph{International Conference on Computer Vision}, pages 1--8.
  IEEE, 2007{\natexlab{b}}.

\bibitem[Yang et~al.(2015)Yang, Gao, and Li]{yang2015efficient}
Kai-Fu Yang, Shao-Bing Gao, and Yong-Jie Li.
\newblock Efficient illuminant estimation for color constancy using grey
  pixels.
\newblock In \emph{IEEE Conference on Computer Vision and Pattern Recognition},
  pages 2254--2263, 2015.

\end{thebibliography}
\end{document}